\documentclass{article}
\usepackage{spconf,amsmath,graphicx,hyperref}
\hypersetup{hidelinks}
\usepackage{booktabs}
\usepackage{caption}
\usepackage{expex}
\usepackage{tipa}
\usepackage{multirow}

\usepackage{siunitx}

\usepackage{xcolor}

\title{ERROR RATE IS NOT ENOUGH: COSTS OF AUGMENTATION IN LOW-RESOURCE ASR}

\name{Katsumi Ibaraki \qquad David Chiang}
    \address{University of Notre Dame \\
            Dept.~of Computer Science and Engineering \\
            \texttt{\{kibaraki, dchiang\}@nd.edu} }

\begin{document}
\ninept
\maketitle
\begin{abstract}
Data augmentation is standard practice for low-resource automatic speech recognition (ASR), but its benefits are almost always reported as aggregate error rates, leaving open \emph{why} and \emph{when} a method helps.
On four low-resource languages and English at five data sizes, we cross two augmentations that modify existing audio (speed perturbation and SpecAugment) with two that modify text transcriptions and synthesize audio from it. We analyze errors at the phoneme level and find that they help for complementary reasons.
SpecAugment has the most reliable aggregate improvements, but increases tone and palatalization errors, whereas text-based methods introduce consonant insertions; in combination, each reduces the errors the other introduces.
This complementarity is predictable. 
Tone and palatalization are each signaled by a single acoustic property and occur least frequently in training (lowest training exposure), and text-based methods that flatten word frequency raise their frequency by 37--38\%, reversing those errors while further improving aggregate error.
The improvement from combining the methods exceeds the sum of their individual improvements; the superadditivity vanishes by 100 hours, limiting the complementarity to the low-resource regime.
Where specific error types matter, 
we recommend selecting augmentation by training exposure and acoustic properties, both measurable before any training.

\end{abstract}
\section{Introduction}

For some endangered and under-documented languages, transcribed speech can be measured in minutes rather than thousands of hours. 
Data augmentation is the standard response, and both modification of existing audio \cite{ko15_interspeech, park19e_interspeech} and conversion of text to new audio \cite{zhong22_interspeech, 9053139, 10889894} are reported to help.
However, when methods are compared by a single aggregate error rate, it establishes \emph{that} they help without establishing \emph{why} or \emph{when} they help.
For language documentation, error \emph{type} matters as much as error rate \cite{adams-etal-2018-evaluation}.
For a linguist that studies (say) tone, an augmentation method that lowers overall error while degrading tone marking may be worse than nothing.
In this paper, we look more deeply into the types of errors that different augmentation methods make.
We find that the method that delivers the best aggregate error rate, SpecAugment \cite{park19e_interspeech}, actually increases some error types, namely, tone and palatalization.
Yet these increases are reversible while still lowering aggregate error; making rare words more frequent in training reverses them, because tone and palatalization sit disproportionately in rare words.

We fine-tune wav2vec2-large-xlsr-53 \cite{conneau2020unsupervised} on four 
low-resource languages and on LibriSpeech subsets from one to one hundred hours, crossing two modification-based methods with two text-based methods---to our knowledge, the first per-category account of what augmentation does to a phoneme inventory.
We make three contributions:
 
\begin{itemize}
\itemsep0em

\item While SpecAugment has the most reliable improvements in WER/CER of any method tested, it increases errors in recognizing tone and palatalization. A text-based augmentation, Random Replacement, reverses these errors while further improving WER/CER.

\item These error categories are predictable: 
SpecAugment increases errors on categories that are rare in training and carried by a single acoustic property, because these categories are more easily masked out. 
Random Replacement, which replaces words uniformly over types, increases the word frequencies of rare categories and reduces errors.

\item A frequency-preserving replacement supports the mechanism: Tag-Based Replacement, which leaves category frequencies unchanged, fails to reverse the tone errors---with SpecAugment, it instead significantly \emph{increases} them (+11\%). 

\end{itemize}

\section{Related Work}

Standard text augmentation relies on external lexical resources such as WordNet \cite{miller_wordnet, wei-zou-2019-eda} or predefined semantic frames \cite{zevallos22_interspeech, hou-etal-2018-sequence}, often unavailable for low-resource languages.
Audio synthesis via TTS \cite{zhong22_interspeech, 9053139, 10889894} usually assumes unpaired text; we follow Robinson et al.\ \cite{robinson22_interspeech} by using a high-resource \emph{pivot} TTS system instead of an in-language TTS model \cite{bartelds-etal-2023-making}.
While Rossenbach et al.\ \cite{rossenbach2020generating} found TTS and SpecAugment contribute largely independently, we revisit this under low-resource constraints.
SpecAugment \cite{park19e_interspeech} and speed perturbation \cite{ko15_interspeech} are the \emph{de facto} acoustic standards, as they require no external data; speed perturbation resamples the waveform and therefore shifts F0.
The IARPA BABEL program \cite{ragni14_interspeech, wrro152840} targeted 10--80 hours; we target 1--5 hours, where glossaries and audio sufficient for pseudo-labeling \cite{lugosch2022pseudolabelingmassivelymultilingualspeech, bhogale2024empowering} are unavailable.
Boundary-aware masking \cite{sym17091478} keeps masks inside phoneme boundaries to protect tonal contours, but requires forced alignments our setting lacks.

Across this literature, augmentation is evaluated almost exclusively by aggregate error rate, so the phonological effects each method introduces go unmeasured.
Akavarapu et al.\ \cite{akavarapu2026hard} relate phoneme accuracy to training frequency across models; augmentation, however, alters the training distribution itself, and its per-phoneme effects remain unexamined.

\section{Method}
\subsection{Augmentation Methods}
We select two methods that modify existing audio and two that generate new text--audio pairs, chosen to differ maximally in what they do to the signal.
The text-based methods extend template-based substitution \cite{zevallos22_interspeech, hou-etal-2018-sequence} to a setting where no external text resources exist; they draw only on the annotations already present in the corpus, and synthesize the result through TTS.

\subsubsection{Text Modification}

\textbf{Tag-Based Replacement} treats the sequence of annotation tags (interlinear glosses, or POS and dependency tags for English) as a fixed template, replacing each word with one sampled from the training word types with the same tag.
It does not guarantee semantic coherence, but it preserves the tag sequence, and with it most of the sentence's grammatical structure.
\textbf{Random Replacement}, a more extreme variant, replaces each position with a word sampled from the training vocabulary; all words remain in-vocabulary, but sentences need not be grammatical.
The replacement ratio $r$ (proportion of positions substituted) is tuned per language and method over $\{0.0, 0.2, 0.4, 0.6, 0.8, 1.0\}$. 
At $r=0.0$, it re-synthesizes the original text, testing whether the benefit comes from novel text or merely additional audio.
Text is synthesized with Kokoro,\footnote{\url{https://huggingface.co/hexgrad/Kokoro-82M}} an open-weight TTS accepting IPA and orthographic input, using five fixed voices across all languages; per-speaker audio was insufficient for voice cloning, and voice conversion \cite{baas2022voice} did not help.

\subsubsection{Audio Modification}

\textbf{Speed perturbation} \cite{ko15_interspeech} applies warp factors in $[0.9, 1.1]$; application probability was tuned from 10--50\%, with 30\% selected for all languages.
\textbf{SpecAugment} \cite{park19e_interspeech} applies time and frequency masking to the feature encoder outputs following \cite{10.5555/3495724.3496768}.
We tuned the mask length between 3 and 10 and the masking probability between 5\% and 50\% by grid search on the validation sets; a length of 5 and a probability of 30\% were chosen uniformly for all languages.

\subsection{Fine-tuning Wav2Vec2}
\label{subsec: finetuning wav2vec2}
We fine-tune the pre-trained wav2vec2-large-xlsr-53 model \cite{conneau2020unsupervised} using a Connectionist Temporal Classification (CTC) loss \cite{10.1145/1143844.1143891}. 
We use a learning rate of $1.0 \times 10^{-4}$ and 
tune training epochs for each language over $\{5$, $10$, $15$, $20$, $30$, $40$, $45$, $50$, $55$, $60$, $80$, $100\}$.
To optimize the data augmentation pipeline, we tuned the ratio of original to synthetic training data for each language across 1:1, 1:2, 1:3, 1:4, and 1:5.  
The selected hyperparameters, optimized on the validation set, are detailed in Table~\ref{tab:metadata}. 
To mitigate tag sparsity in our Tag-Based Replacement, we relaxed the exact-match constraint, falling back to the nearest tag by edit distance so a replacement is always found and the target ratio is met.
Model training was conducted on a single NVIDIA A10 GPU, with runtimes ranging from 3 to 160 hours depending on the dataset size and augmentation ratio.

\section{Experiment Setup}

\begin{table}[t]
\centering\scriptsize
\setlength{\tabcolsep}{4pt}
\caption{
Corpus statistics and selected hyperparameters: minutes of transcribed
audio, speakers, training word tokens and types, tag types, replacement
ratio $r$ and original:synthetic mix (Tag/Random), and epochs.
LibriSpeech (LS) splits are subsets of \texttt{train-clean-100}
\cite{panayotov2015librispeech}.
}
\begin{tabular}{@{}lrrrrrccr@{}}
\toprule
 & & & \multicolumn{2}{c}{Train Words} & Tag & Repl.\ $r$ & Mix & \\
 \cmidrule(lr){4-5}
 & Min. & Spk. & Tok. & Typ. & Typ. & (T/R) & (T/R) & Ep. \\
\midrule
Vatlongos & 286 & 60 & 23315 & 2823 & 422 & .2/.6 & 1:3/1:3 & 40 \\
Nashta & 48 & 4 & 3982 & 886 & 637 & .6/.2 & 1:2/1:1 & 60 \\
S.\ Buryat & 97 & 4 & 7895 & 2276 & 2242 & .4/.2 & 1:3/1:1 & 45 \\
Kakabe & 99 & 34 & 12512 & 1525 & 259 & .2/1.0 & 1:2/1:3 & 60 \\
\midrule
\multicolumn{9}{@{}l}{English (LibriSpeech)} \\
\quad LS-1h & 60 & 169 & 6967 & 2149 & 383 & .2/.4 & 1:1/1:1 & 60 \\
\quad LS-2h & 120 & 222 & 13528 & 3568 & 488 & .2/.4 & 1:1/1:1 & 50 \\
\quad LS-5h & 301 & 249 & 34992 & 6566 & 646 & .2/.4 & 1:1/1:1 & 30 \\
\quad LS-20h & 1207 & 251 & 138790 & 14333 & 888 & .2/.4 & 1:1/1:1 & 10 \\
\quad LS-100h & 6036 & 251 & 693914 & 29598 & 1184 & .2/.4 & 1:1/1:1 & 5 \\
\bottomrule
\end{tabular}
\label{tab:metadata}
\end{table}

\subsection{Data}
We tested on four low-resource languages,
all with five hours or less of audio.
We allowed speaker overlap between splits following \cite{WEI20221, liu-etal-2023-investigating},
and 
as the target application---assistive transcription of a field linguist's unlabeled collection---inherently features such overlap.
While the level of transcription detail varies across our data sources, we treat all IPA-based transcriptions as phonemic representations and refer to the units as \textit{phonemes}. 

Vatlongos (Vanuatu \cite{ridge2018language}),
Nashta (a South Slavic variety of the Balkans \cite{adamou2022nashta}), Shinekhen Buryat (Mongolic, Inner Mongolia \cite{yamakoshi2011three}) and Kakabe (Mande, Guinea \cite{vydrina2017corpus}) are drawn from the Pangloss Collection and the Tokyo University of Foreign Studies repository. All four are annotated with IPA transcriptions, translations, and interlinear glosses; an example from Shinekhen Buryat is shown in (1).
Kakabe marks tone, Nashta marks stress, and Nashta and Shinekhen Buryat mark palatalization.
LibriSpeech is a corpus of read English speech \cite{panayotov2015librispeech}, to which we add \texttt{spaCy} \cite{honnibal2020spacy} POS and syntactic dependency tags, emulating a documentation-style annotation.

\pex[exskip=0.3ex, aboveglftskip=0.5ex,extraglskip=0ex]<withparts> 
\begingl 
\gla \textipa{iimerxuu}	\textipa{buleg}	\textipa{buleg-eer-ee}	\textipa{bai-dag}//
\glb suchlike	group	group-\textsc{ins-refl}	be-\textsc{ptcp.hbt} //
\glft `They lived in groups like this.'//
\endgl
\xe

\subsection{Evaluation}
\label{subsec:eval}

We report Word Error Rate (WER) with Phoneme Error Rate (PER) for the four IPA-transcribed low-resource corpora and Character Error Rate (CER) for orthographic English.
Significance is assessed against the baseline with paired bootstrap resampling \cite{bisani_bootstrap, koehn-2004-statistical}, 10{,}000 samples at $\alpha = 0.05$; the same resampling is applied to per-category error counts in Table~\ref{tab:error_analysis}, with a Holm--Bonferroni correction \cite{holm1979simple} across categories.
Because the repair claims of Section~\ref{subsec:complementarity} compare SpecAugment alone with SpecAugment plus a text-based method, rather than either with the baseline, those are additionally tested head-to-head.
To distinguish which errors a method introduces, we align each hypothesis to its reference at the phoneme level using Levenshtein distance, and we label every substitution, deletion, and insertion with a phonological category (those of Table~\ref{tab:error_analysis}). 

Finally, we rank every individual unit in a corpus (each vowel, consonant, tone, stress, and palatalized segment) by its training token count and report the median rank of each category's members as a percentile of the full inventory, which we call the \emph{training
exposure percentile}. 
This makes exposure comparable across corpora of different sizes and inventories. 
We also report raw token counts where magnitude matters.

\begin{table}[t]
\centering

\scriptsize
\caption{Results (\%). $^*$: significant improvement over the baseline at $\alpha=.05$, paired bootstrap resampling \cite{bisani_bootstrap, koehn-2004-statistical}; bold is best before rounding. SP = speed perturbation, SA = SpecAugment.}
\setlength{\tabcolsep}{4.7pt}
\begin{tabular}{
@{}
l
S[table-format=2.1, table-space-text-post={*}, table-align-text-post=false]
S[table-format=2.1, table-space-text-post={*}, table-align-text-post=false]
S[table-format=2.1, table-space-text-post={*}, table-align-text-post=false]
S[table-format=2.1, table-space-text-post={*}, table-align-text-post=false]
S[table-format=2.1, table-space-text-post={*}, table-align-text-post=false]
S[table-format=2.1, table-space-text-post={*}, table-align-text-post=false]
S[table-format=2.1, table-space-text-post={*}, table-align-text-post=false]
S[table-format=2.1, table-space-text-post={*}, table-align-text-post=false]
@{}
}
\toprule


& \multicolumn{2}{c}{Vatlongos}
& \multicolumn{2}{c}{Nashta}
& \multicolumn{2}{c}{S. Buryat}
& \multicolumn{2}{c}{Kakabe} \\

\cmidrule(lr){2-3}
\cmidrule(lr){4-5}
\cmidrule(lr){6-7}
\cmidrule(lr){8-9}

\textbf{Method}
& {PER} & {WER}
& {PER} & {WER}
& {PER} & {WER}
& {PER} & {WER}

\\
\midrule

Base
& 13.4 & 36.8
& 18.4 & 59.1
& 14.5 & 45.3
& 23.6 & 47.8 \\

\quad + {SP}
& 13.1 & 36.5
& 18.5 & 60.4
& 14.6 & 46.7
& 23.3 & 46.7 \\

\quad + SA
& 12.3$^*$ & 34.7$^*$
& 17.6$^*$& 56.0$^*$
& 13.3$^*$& 44.4
& 21.6$^*$ & 45.7$^*$ \\



\midrule

Tag
& 12.6 & 34.8
& 18.1 & 56.5
& 14.3 & 45.2
&24.2 & 47.3 \\

\quad + SP
& 13.0 & 35.9
& 18.7 & 58.9
& 13.9 & 44.2
&23.9 & 47.9 \\

\quad + SA
& 11.5$^*$& 31.9$^*$
& 16.5$^*$& 53.2$^*$
& 13.4$^*$& 43.5
&22.3$^*$ & 46.3 \\

\midrule

Rand.
& 12.9 & 36.0
& 18.6 & 57.6
& 14.5 & 45.2
&23.5 & 46.5 \\

\quad + SP
& 12.8 & 35.4
& 18.0 & 55.4
& 14.2 & 44.7
&23.4 & 46.9 \\

\quad + SA
& \bfseries 11.5$^*$
& \bfseries 31.6$^*$
& \bfseries 16.1$^*$
& \bfseries 51.1$^*$
& \bfseries 13.2$^*$
& \bfseries 42.9
& \bfseries 21.2$^*$  
& \bfseries 43.1$^*$ \\

\bottomrule
\end{tabular}

\smallskip

\setlength{\tabcolsep}{2.3pt}
\begin{tabular}{
@{}
l
S[table-format=2.1, table-space-text-post={*}, table-align-text-post=false]
S[table-format=2.1, table-space-text-post={*}, table-align-text-post=false]
S[table-format=2.1, table-space-text-post={*}, table-align-text-post=false]
S[table-format=2.1, table-space-text-post={*}, table-align-text-post=false]
S[table-format=2.1, table-space-text-post={*}, table-align-text-post=false]
S[table-format=2.1, table-space-text-post={*}, table-align-text-post=false]
S[table-format=2.1, table-space-text-post={*}, table-align-text-post=false]
S[table-format=2.1, table-space-text-post={*}, table-align-text-post=false]
S[table-format=2.1, table-space-text-post={*}, table-align-text-post=false]
S[table-format=2.1, table-space-text-post={*}, table-align-text-post=false]
@{}
}
\toprule

& \multicolumn{10}{c}{{English (LibriSpeech)}} \\

& \multicolumn{2}{c}{1h}
& \multicolumn{2}{c}{2h}
& \multicolumn{2}{c}{5h}
& \multicolumn{2}{c}{20h}
& \multicolumn{2}{c}{100h} \\

\cmidrule(lr){2-3}
\cmidrule(lr){4-5}
\cmidrule(lr){6-7}
\cmidrule(lr){8-9}
\cmidrule(lr){10-11}

\textbf{Method}
& {CER} & {WER}
& {CER} & {WER}
& {CER} & {WER}
& {CER} & {WER}
& {CER} & {WER}

\\
\midrule

Base
& 6.5 & 24.4
& 4.8 & 17.3
& 2.9 & 10.1
& 3.0 & 9.0 
& 1.9 & 6.0\\

\quad + SP
& 6.0$^*$& 22.0$^*$
& 4.1$^*$& 14.9$^*$
& 3.0 & 10.4
& 2.4$^*$& 7.9$^*$
& 1.9 & 5.9\\

\quad + SA
& 5.0$^*$& 18.7$^*$
& 4.4$^*$& 16.5$^*$
& 3.0 & 10.9
& 2.0$^*$& 7.0$^*$
& 1.6$^*$& 5.5$^*$\\



\midrule

Tag
& 4.3$^*$& 15.0$^*$
& 3.7$^*$& 12.6$^*$
& 2.7$^*$& 8.8$^*$
& 2.0$^*$& 6.6$^*$
& 1.9 & 5.7\\

\quad + SP
& 4.4$^*$& 15.4$^*$
& 3.7$^*$& 12.7$^*$
& 2.7$^*$& 8.9$^*$
& 2.1$^*$& 6.6$^*$
& 2.1 & 6.3\\

\quad + SA
& 4.1$^*$& 14.9$^*$
& \bfseries 3.4$^*$& \bfseries 11.9$^*$
& \bfseries 2.3$^*$& \bfseries 7.8$^*$
& \bfseries 1.7$^*$& \bfseries 5.8$^*$
& \bfseries 1.4$^*$& \bfseries 4.9$^*$\\

\midrule

Rand.
& 4.7$^*$& 16.4$^*$
& 3.9$^*$& 13.1$^*$
& 3.0 & 9.7$^*$
& 2.5$^*$& 7.6$^*$
& 2.0 & 6.0\\

\quad + SP
& 4.8$^*$& 16.8$^*$
& 3.8$^*$& 12.9$^*$
& 2.9 & 9.6$^*$
& 2.7$^*$& 8.2$^*$
& 2.0 & 5.9\\

\quad + SA
& \bfseries 4.0$^*$
& \bfseries 13.9$^*$
& 3.5$^*$
& 12.2$^*$
& 2.6$^*$
& 8.7$^*$
& 1.9$^*$
& 6.6$^*$
& 1.6$^*$& 5.5$^*$\\

\bottomrule

\end{tabular}
\label{tab:combined-results}
\end{table}

\newcommand{\sig}[1]{\textbf{#1}}
\newcommand{\sigd}[1]{\textbf{#1}\rlap{\textsuperscript{\dag}}}

\begin{table*}[htbp]
\centering
\caption{Relative change (\%) in error count vs.\ the baseline; negative is improvement. Segmental rows pool the four languages; tone is Kakabe-only, stress Nashta-only, palatalization Nashta and Shinekhen Buryat. 
Tone, stress, and palatalization rows pool substitution (sub), insertion (ins), and deletion (del) channels; a mark on an inserted or deleted segment counts in both its segmental and non-segmental row, so rows overlap and Total sums the listed channels. \textbf{Exp.}\ is the median training exposure percentile (Section~\ref{subsec:eval}); rows are ordered by it, with the rule separating high from low. \textbf{Bold}: $p<.05$, paired bootstrap; \dag: survives Holm correction across categories. SA = SpecAugment, SP = speed perturbation.}
\label{tab:error_analysis}
\scriptsize
\begin{tabular}{llrr rr rrr rrr}
\toprule
& & & & & & \multicolumn{3}{c}{\textbf{Tag}} & \multicolumn{3}{c}{\textbf{Rand}} \\
\cmidrule(lr){7-9} \cmidrule(lr){10-12}
\textbf{Error Type} & \textbf{Kind} & \textbf{Exp.} & \textbf{Errors} & \textbf{SA} & \textbf{SP} & \textbf{Base} & \textbf{+SA} & \textbf{+SP} & \textbf{Base} & \textbf{+SA} & \textbf{+SP} \\
\midrule
stress & suprasegmental & 98 & 79 & -5 & +15 & +10 & +22 & +15 & +6 & +4 & +3 \\
sub: vowel & segmental & 89 & 981 & -6 & +2 & +4 & -6 & +1 & +0 & \sigd{-10} & -4 \\
ins: vowel & segmental & 89 & 961 & \sigd{-14} & \sigd{-11} & \sig{-7} & \sigd{-15} & -4 & +4 & \sigd{-10} & -0 \\
del: vowel & segmental & 89 & 870 & +1 & +5 & +6 & -2 & +5 & +1 & -4 & +3 \\
sub: consonant & segmental & 51 & 762 & \sigd{-15} & +1 & +3 & \sigd{-11} & -1 & +2 & \sigd{-17} & -0 \\
ins: consonant & segmental & 51 & 865 & \sig{-10} & -6 & +3 & \sigd{-13} & +3 & \sigd{+18} & \sigd{-11} & -0 \\
del: consonant & segmental & 51 & 843 & \sig{-7} & +4 & -0 & \sigd{-11} & +1 & +1 & \sigd{-14} & -1 \\
sub: vowel$\leftrightarrow$consonant & segmental & -- & 673 & \sigd{-18} & +5 & -1 & \sigd{-21} & +3 & +4 & \sigd{-15} & +1 \\
\midrule
tone & suprasegmental & 26 & 488 & +4 & +3 & +2 & \sig{+11} & +3 & -2 & -5 & +0 \\
palatalization & diacritic & 23 & 69 & +7 & -3 & -14 & -19 & -12 & -9 & -14 & -6 \\
\midrule
Total & & & 6591 & \sig{-8} & +0 & +1 & \sig{-9} & +1 & \sig{+4} & \sig{-11} & -0 \\
\bottomrule
\end{tabular}
\end{table*}


\section{Results and Discussion}

\subsection{Overall Performance}
 
In Table~\ref{tab:combined-results}, no text-based method alone reaches significance on either metric across the four low-resource languages; the largest single improvement is 2.6\% WER for Tag-Based Replacement on Nashta.
$r = 0.0$ was never selected: re-synthesized audio of the original text never beat novel text, so the value of text-based augmentation lies in the text, not merely in additional audio.
Speed perturbation is also not significant in any of its twelve comparisons.
SpecAugment is the exception, improving PER and WER in all combinations by an average of 1.5\% PER and 2.8\% WER absolute; a practitioner testing methods individually against a baseline would adopt it, as the only single method that improves both metrics in every low-resource comparison.

Combinations change the picture.
Random Replacement with SpecAugment gives the lowest PER and WER in all four languages, reducing WER by 2.4--8.0\% absolute (5.3--14.1\% relative).
On Nashta, Random Replacement alone improves WER by 1.5\% and SpecAugment alone by 3.1\%, yet the combination improves it by 8.0\%.
This super-additivity holds in all four languages, but vanishes by 100\,h in LibriSpeech, where the methods contribute largely independently,
matching findings of Rossenbach et al. \cite{rossenbach2020generating}.

\subsection{Phoneme-level Analysis}
\label{sec:analysis}
Analyzing by phonological category shows that while SpecAugment decreases overall errors, it causes some error types to increase.
On Kakabe, SpecAugment alone improves both PER and WER significantly and is the strongest single method by either metric, but increases tone errors by 4\%; Random Replacement, by itself significant on neither metric, reduces them by 2\%.
Neither change is significant against the baseline on its own, but adding Random Replacement to SpecAugment reduces tone errors significantly relative to SpecAugment alone: by 9\% ($p = .043$; Section~\ref{subsec:complementarity}).
Aggregate error rate shows none of this: neither the tone errors SpecAugment adds nor their reversal by Random Replacement.

See Table~\ref{tab:error_analysis}.
SpecAugment's improvements are strongest in vowel and consonant errors (\textminus18\%),
while the categories in which SpecAugment appreciably increases errors are tone (+4\%) and palatalization (+7\%).
Neither reaches significance on its own, but the tone increase does reach significance in one combination, Tag-Based Replacement with SpecAugment (+11\%, 95\% CI $[+2, +22]$, $p = .025$), for reasons we return to in Section~\ref{subsec:complementarity}.

One possible explanation is \textbf{exposure}, or how often these categories occur. The vowel and consonant categories are among the most frequent, while tone and palatalization are the two scarcest categories.

A second possible explanation is what we call \textbf{signaling}, or how many different kinds of acoustic cues a category has. 
When SpecAugment masks out one out of several cues, the model must learn to combine evidence from the others. 
Vowels and consonants differ in periodicity, duration, and gross
spectral shape, so masking one of these leaves the others intact.
Stress is typically signaled by several acoustic properties, including duration, intensity, and F0, and improves slightly.
By contrast, both tone and palatalization are signaled by essentially one acoustic property: tone by F0, and palatalization by formant transitions, unlike the segmental categories above.

A third possible reason is that tone and palatalization are both non-segmental (that is, they are realized on or across vowels and consonants).
But stress is also non-segmental, and it improves under SpecAugment.
The difference is that tone and palatalization have low exposure and single-cue signaling, while stress has neither: it sits at the 98th exposure percentile, appearing in 87\% of Nashta word types, and is carried by several cues.
So exposure and signaling, not segmentality, are the more likely reasons SpecAugment increases tone and palatalization errors.

An alternative explanation of the increase in tone errors would be mask size: a mask longer than the tone-bearing segment hides the whole tone, so longer masks should hurt tone more.
To test this, we swept the mask length on Kakabe from 60 to 200\,ms. Aggregate performance peaked at 100\,ms (PER 22.7, 21.6, 22.2, 22.5), but tone errors increased at every width and did not grow with width (+9\%, +5\%, +1\%, +5\%). So mask size does not explain the increase.

Speed perturbation reduces insertions (\textminus11\% vowels, \textminus6\% consonants) but increases deletions and substitutions, producing a near-zero net effect.
The errors it introduces fall on the dimensions it manipulates---it shifts F0, which carries tone, and it warps duration, affecting stress and consonants.

 
\subsection{Method Complementarity}
\label{subsec:complementarity}

The combinations' advantage runs in two directions:
each method repairs the errors the other introduces.
Random Replacement raises consonant insertions by 18\% while leaving deletions largely unchanged; the model outputs too many symbols, and its total labeled errors rise significantly (+4\%, $p = .004$)---yet it forms the best-performing combination.
One possible explanation is that in synthetic audio, every frame is a clean phoneme or silence---synthetic speech measurably differs in distribution from real recordings \cite{minixhofer23_interspeech}---so the model learns no mapping from real-world non-speech frames (breath, noise) to blank, and inserts symbols instead.
SpecAugment fixes this.
Masking contiguous regions prevents the model from depending on any single local feature and pushes it to integrate longer context \cite{park19e_interspeech, vieting25_interspeech}, suppressing exactly these locally-triggered insertions.
If SpecAugment worked by simply making the model output fewer symbols overall, deletions should rise as insertions fall; instead, consonant deletions drop 11\% and 14\% when Tag-Based and Random Replacement are combined with SpecAugment, beyond SpecAugment's own 7\%.
Insertions and deletions fall together, so masking makes emissions more accurate, not just rarer.

The repair in the other direction---Random Replacement reducing the errors masking introduces---is not coincidental.
SpecAugment's errors increase precisely where exposure is lowest.
Tone in Kakabe and palatalization in Nashta and Shinekhen Buryat all fall below the 30th training exposure
percentile (25.7th, 18.4th and 27.0th), while vowels fall above the 80th (82.9th, 89.8th, 89.2nd); an individual Kakabe tone value occurs in a median of 81 training tokens against 2{,}120 for an individual vowel.
Random Replacement draws uniformly from the training vocabulary, so its synthetic text is flatter in word frequency than the real corpus, and tone and palatalization, concentrated in less frequent words, gain token share.
In Kakabe, flattening raises the tone-bearing token share by 38\% (palatalized tokens by 37\% in Nashta), while vowel and consonant shares move by at most 7\%.

Tag-Based Replacement also samples word types uniformly, but only among words sharing a tag (or a close tag, under the fallback of Section~\ref{subsec: finetuning wav2vec2}).
The sequence of tags itself is kept, with its natural frequencies, and since tone and palatalization are in some tag classes and not others, their token shares are preserved almost exactly (tone at 1.00, palatalization at 0.97).

If exposure drives the errors, flattening word frequencies should reverse them and tag-based substitution should not.
Tested directly against SpecAugment rather than the baseline, adding Random Replacement reduces tone errors by 9\% (95\% CI $[-16, -0.4]$, $p = .043$) and palatalization errors by 20\% ($p = .061$); no segmental category significantly worsens, and the addition significantly improves even the total labeled errors over SpecAugment alone ($-2.7\%$, $p = .013$).
Tag-Based Replacement behaves as predicted.
It leaves exposure unchanged, fails to reverse the tone errors, and with SpecAugment worsens them (+11\% against the baseline, $p = .025$).
One result falls outside the exposure account: Tag-Based Replacement also reduces palatalization errors ($-24\%$, $p = .041$) despite changing exposure by nothing.
Whatever drives this repair, it is not token frequency; identifying the channel is left to future work.

 
\subsection{Limitations}

Training on randomly replaced text exposes the model to invalid syntax, and text-based augmentation does not address out-of-vocabulary words.
English differs from the target languages: read speech, orthographic transcription, and presence in both models' pre-training.
Three links in the mechanism remain untested directly: signaling is inferred from phonetics, low exposure and single-property signaling coincide for tone and palatalization, and Tag-Based Replacement's reduction of palatalization errors, achieved without any exposure change, is not explained by our account.
We mask frames and replace words; doing either at the phoneme level would need boundaries that neither CTC alignment nor an aligner trained on 1--5\,h reliably gives.
The tone results also rest on a single language (Kakabe), and palatalization on two.
Finally, five fixed pivot-TTS voices likely underestimate what text-based augmentation can achieve, since speaker diversity drives synthetic data's value \cite{9053139, 10889894}.

\section{Conclusion}

A method's aggregate improvement says nothing about what it preserves; here the method with the most reliable aggregate improvements is the one that increases tone and palatalization errors, and no method we test is free of such trades.
In the low-resource regime, SpecAugment with Random Replacement is the default: it gave the lowest error rates in all four languages, improving aggregate error over SpecAugment alone while repairing the cost of masking.
Before adopting any augmentation, a practitioner can rank everything the target transcription distinguishes by training exposure and signaling; a masking method will increase errors on what is low on both.
The same inspection suggests the repair: making the rare words that carry these categories more frequent reverses the errors.
Where transcripts serve documentation, augmentation choices should be checked against an analysis of error types, not the total error rate alone.

\section{Acknowledgements}
The authors have no relevant financial or nonfinancial interests to disclose.

\section{Compliance with Ethical Standards}
This research was conducted retrospectively using human subject data made available in open access by the Pangloss Collection 
(Vatlongos\footnote{\url{https://pangloss.cnrs.fr/corpus/Vatlongos}}: \href{https://creativecommons.org/licenses/by-nc-nd/3.0/}{CC BY-NC-ND 3.0}, Nashta\footnote{\url{https://pangloss.cnrs.fr/corpus/Nashta}}: \href{https://creativecommons.org/licenses/by-nc/2.5/}{CC BY-NC 2.5}, Kakabe\footnote{\url{https://pangloss.cnrs.fr/corpus/Kakabe}}: \href{https://creativecommons.org/licenses/by-nc-nd/3.0/}{CC BY-NC-ND 3.0}), Tokyo University of Foreign Studies\footnote{\url{https://tufs.repo.nii.ac.jp/search?search_type=2&q=1729497608274}}
(\href{https://creativecommons.org/licenses/by-sa/4.0/deed.en}{CC BY-SA 4.0}),
and Panayotov et al.~\cite{panayotov2015librispeech} (\href{https://creativecommons.org/licenses/by/4.0/}{CC BY 4.0}). 
Ethical approval was not required as confirmed by the license attached with the open access data.

\begingroup

\makeatletter
\patchcmd{\thebibliography}{\relax}{\footnotesize}{}{}
\makeatother

\bibliographystyle{IEEEbib}
\bibliography{strings,refs} 
\endgroup

\end{document}